\documentclass[11pt]{article}
\usepackage{arxiv}

\input{glyphtounicode}
\usepackage[utf8]{inputenc}
\usepackage[T1]{fontenc}
\usepackage{lmodern}
\usepackage{hyperref}
\usepackage{url}
\usepackage{booktabs}
\usepackage{amsmath}
\usepackage{graphicx}
\usepackage{natbib}
\usepackage{xcolor}
\usepackage[most]{tcolorbox}

\usepackage{fancyvrb}
\makeatletter
\renewcommand{\verbatim@font}{\ttfamily\footnotesize}
\makeatother

\hypersetup{
  colorlinks=true,
  linkcolor=blue!50!black,
  citecolor=blue!50!black,
  urlcolor=blue!50!black
}

\title{Regimes: An Auditable, Held-Out-Gated Improvement Loop\\[2pt] Demonstrated on LongMemEval with ActiveGraph}

\author{
  Yohei Nakajima\thanks{Regimes builds on prior self-modification experiments on the ActiveGraph runtime (selfgraph and behaviordrafts), documented at \url{https://activegraph.ai/blog/self-modification-is-not-binary} and \url{https://activegraph.ai/blog/code-without-authority}.}\\
  Untapped Capital\\
  \texttt{activegraph.ai}
}

\date{June 8, 2026}

\begin{document}
\maketitle

\begin{abstract}
Autonomous improvement loops are hard to trust because the improvement process is usually external scaffolding bolted onto the agent: failures go unlogged, diagnoses cannot be replayed, and each promote-or-discard decision lands in a side database rather than in the agent's own history. We show that an event-sourced agent runtime removes that friction and turns controlled improvement into a first-class workflow. When the agent's state is a deterministic projection of an append-only event log, failures are recorded, a recorded run replays exactly from its log, candidate patches scope to typed pipeline seams, gates are auditable, and every promotion or discard is itself an event. We demonstrate this with Regimes, an improvement loop built on the ActiveGraph runtime. Regimes diagnoses failed evaluations, proposes a repair at a specific point in the agent's pipeline, and promotes that repair only when it passes static checks, sandbox execution, in-sample evaluation, and validation on held-out examples. The loop is target-agnostic: the same control flow runs against different agent tasks through a common interface. ActiveGraph is what makes this auditable rather than anecdotal: the runtime is event-sourced, so a recorded run replays deterministically from its log, model and tool responses are cached for exact replay when the cache is available, and the loop's own history (diagnosis, proposed repair, gate outcomes, promotion or discard) is itself an auditable event log rather than a side database.

We demonstrate the loop on LongMemEval-S, where the dominant failure is not retrieval but reconciliation: the relevant evidence is already present in the assembled context, yet the reader answers incorrectly. Across five seeded held-out splits, Regimes discovers reader-prompt repairs that improve final held-out accuracy by $+0.05$ to $+0.10$ in four splits and by $+0.01$ in one over-promotion split. Two splits are individually significant under a simple paired test (seed 5's value is not adjusted for its four-step sequential promotion structure); the pooled flip count is positive but should be read as descriptive, because the splits come from the same 500-question pool. The detailed statistics (per-split McNemar tests, pooled discordances, and the same-pool caveat) are in the results section rather than here.

The case study shows both the promise and the limits of autonomous improvement. The held-out gate rejects many overfit candidates, including nearly every retrieval-reweighting repair, while the weak split reveals a concrete next improvement: a stricter promotion threshold and a plateau-aware stopping rule. More broadly, the discovered prompt repairs point past themselves: their successes reveal which evidence-use behaviors matter, and their regressions reveal when a behavior should not fire. Durable agent improvements should convert those behaviors into guarded deterministic operators that condition on detected structure rather than on coarse prose rules. The durable contributions are ActiveGraph as an auditable substrate that makes controlled improvement loops tractable, the reproducible held-out-gated loop it supports, the failure-regime taxonomy that routes each failure to a pipeline location (whose marginal value over an unrouted baseline is the primary open question), and the prompt-as-discovery-probe hypothesis.
\end{abstract}

\section{Introduction}

Long-running agents fail in repeated, diagnosable ways. As an agent accumulates history, retrieves over it, assembles a context, and reads that context to answer, each stage can break, and the same kinds of breakage recur. What is missing is not another point fix but a loop: something that can detect a recurring failure, propose a repair, test the repair safely, and keep it only if it works on examples it was not tuned on. This paper presents such a loop, called Regimes, and uses one agent task as a detailed case study.

Two plain-language commitments shape the design. First, a repair is kept only if it works on held-out examples, not just on the examples used to write it. We call this a held-out-gated promotion, and it is what separates a real improvement from a candidate that merely overfits the cases the loop happened to see. Second, the same loop should run against different agent tasks without rewriting its core. We call this target-agnostic: the diagnosis, gating, and rotation logic talk to any task through a common interface, and only the task-specific pieces (how to evaluate it, which parts of its pipeline may be edited, how to label its failures) are swapped in.

ActiveGraph is part of the core story, not a deployment detail. The loop runs on the ActiveGraph runtime \citep{nakajima2026log}, an event-sourced system whose state is a deterministic projection of an append-only event log, with a cache that records model and tool responses. This is what makes autonomous improvement auditable rather than anecdotal: a run replays exactly from its log without new model calls, the evaluation is controlled and repeatable, and the loop's own history (each diagnosis, each proposed repair, each gate outcome, each promotion or discard) is itself an auditable event log rather than a side database. A self-improvement claim is only as trustworthy as the substrate that records it, and an event-sourced substrate lets a reader audit every step. This is where the substrate, not the algorithm, is the lever.

The case study is LongMemEval-S \citep{wu2024longmemeval}, a long-context memory benchmark. A prior analysis on the same substrate found that deterministic retrieval already gets the right evidence into context most of the time, so the dominant remaining failure is not retrieval but reconciliation: the relevant evidence is present in the assembled context, and the reader still answers incorrectly. We give this failure a short name for reuse, assemble-internal, meaning the evidence was assembled into context but used incorrectly. Re-ranking retrieved turns cannot fix it, because the turns are already in context; only a change to how the reader uses the evidence can.

The loop has a small vocabulary, defined here once and used consistently after. A failure regime is the diagnosed reason a question failed (for example, evidence dropped at the context budget, or assemble-internal). An action seam is the part of the pipeline the loop is allowed to edit in response (re-weight retrieval scores, reorder the assembled turns, or edit the reader's prompt). The regime-to-seam mapping is the fixed routing from a diagnosed regime to the seam that can address it. So the loop, in one sentence: diagnose the dominant failure regime, route it to its seam, have a language model author a repair at that seam, and promote the repair only if it survives static checks, sandbox execution, in-sample evaluation, and held-out validation. Repairs are executable patches over named pipeline points; for the reconciliation case study the seam that matters edits the reader-prompt fragment, so the repairs that work here are reader-prompt edits, not arbitrary generated runtime code.

One idea runs through the whole paper and is worth stating plainly up front: the promoted prompt is not the destination. A prompt repair is a high-bandwidth probe. When independently discovered repairs keep encoding the same evidence-use behavior, and their occasional regressions reveal exactly when that behavior should not fire, the prompt has pointed at a candidate deterministic operator. The lasting value of the loop is less the prompt it promotes than the operator that prompt reveals. This reframes a modest accuracy gain as a discovery instrument, and it connects the case-study result (Section 5), the failure analysis (Section 5.9), and the forward agenda (Section 8). It is also why the substrate matters past the experiments here: the operators these probes reveal belong as typed projections on the event log itself (Section 8.4).

Regimes sits at the intersection of four maturing threads: self-improving agents (SICA, \citet{robeyns2025sica}; GRASP, \citet{moll2026grasp}), held-out prompt and program optimization (DSPy, \citet{khattab2023dspy}), executable failure-mode analysis (MAST, \citet{cemri2025mast}; AgentDebug, \citet{zhu2025agentdebug}), and long-context memory benchmarks (LongMemEval and LongMemEval-V2; \citealp{wu2024longmemeval}; \citealp{wu2026longmemeval2}).

The contributions, ordered from the enabling substrate to the specific case study:
\begin{enumerate}
\item ActiveGraph as an auditable substrate for controlled improvement loops: event sourcing, deterministic replay from the log, cached model and tool responses, and the loop's own history recorded as events are what make autonomous improvement easy to build, replay, inspect, constrain, and audit. The empirical sections demonstrate that this affordance is real: the loop re-homes cleanly across tasks (Section 4) and produces a held-out-validated, fully auditable improvement run (Sections 5, 6).
\item Regimes as one improvement loop built on that substrate: diagnose, propose, gate, promote, rotate, with the full history recorded as events, demonstrated to improve one task (LongMemEval-S) under one reader. Its target-agnosticism is established at the level of interface and control flow, not as a multi-task empirical result.
\item A held-out promotion gate that rejects overfit repairs, installed before the action space was widened to free-text prompt edits. The gate is the loop's trust guardrail and the component it shares with prior gated optimizers (DSPy, GRASP) rather than its novelty: one run's in-sample +0.18 collapsed to held-out +0.04, and the gate kept the overfit portion out.
\item A failure-regime taxonomy that routes each failure to the pipeline location that can address it (score, assembly, or reader-prompt). The routing is the organizing heuristic of the loop; whether the diagnosis step adds value beyond supplying failed examples to a held-out-gated author is supported here only indirectly and is the primary open question (Section 5.8, Threat 11).
\item A LongMemEval-S case study showing modest but consistent improvement: held-out gains of +0.05 to +0.10 in four of five seeded splits, near zero in the fifth, with the fifth split itself an informative over-promotion finding.
\item A hypothesis, induced from the discovered repairs and two analyzed regressions, that prompt repairs are useful discovery probes but should later become guarded deterministic operators that fire on detected structure rather than coarse prose rules. No operator is built or evaluated here; this is a forward bet, not a result.
\end{enumerate}

The substrate-supported loop is target-agnostic at the level of interface and control flow, demonstrated by re-homing the loop onto a second, structurally different task (text-to-SQL) with byte-identical loop behavior. We do not claim that the loop improves multiple tasks empirically; the measured improvement is specific to LongMemEval-S under a single reader (claude-sonnet-4-6).

The calibrated claim, stated once so it frames what follows: an event-sourced runtime makes the diagnose-route-repair loop tractable and auditable, and on LongMemEval-S that loop produces a modest, directionally consistent held-out improvement, with two of five splits individually significant and the pooled count significant only as a descriptive same-pool summary. The lasting contribution is the substrate-supported machinery plus the evidence-use operators its prompt repairs reveal, not a new record on the benchmark.

Reader's map. The paper has one spine: an event-sourced runtime makes controlled improvement loops tractable, Regimes is one such loop, LongMemEval is the stress test, and guarded operators are the implication. Section 2 places the work among self-improving agents, prompt optimization, and failure taxonomies. Section 3 defines the loop and its vocabulary. Section 4 tests whether the loop is genuinely target-agnostic by re-homing it onto a second task. Section 5 reports the LongMemEval case study, including the multi-seed replication and the seed-101 over-promotion finding. Section 6 audits the measurement itself, including the regime classifier and the determinism claim. Section 7 lists threats. Section 8 argues why the promoted prompts point past themselves to guarded operators, sketched as proposed designs. Sections 9 and 10 give design lessons and conclude.

\section{Background and related work}

Self-improving and gated loops. SICA \citep{robeyns2025sica} demonstrates an LLM coding agent that edits its own codebase to improve benchmark performance, removing the meta-agent and target-agent split. Regimes shows the improvement loop can be made target-agnostic while preserving deterministic replay on an event-sourced substrate. Reflexion \citep{shinn2023reflexion} is the foundational verbal-reflection precursor, in which an agent writes natural-language critiques of failed trajectories and conditions on them in later attempts. ExpeL \citep{zhao2024expel} extends verbal feedback to cross-trajectory rule extraction.

The closest concurrent peer is GRASP \citep{moll2026grasp}, and the comparison is worth drawing precisely, because it is where this paper's contribution is most easily misread. GRASP's center of gravity is an algorithmic self-improvement method: propose natural-language skills, gate them against held-out performance, and keep the ones that generalize. Regimes' center of gravity is a runtime story: if an agent is built on an event-sourced graph, then improvement loops become natural to build, replay, scope, and audit, and Regimes is one such loop. The two genuinely overlap on a single finding, arrived at independently: the validation gate, not the skill or transform writer, is the component that makes the result real. GRASP shows this by an ablation where removing the gate collapses accuracy to baseline; Regimes shows it by the held-out gate discarding overfit transforms and discounting overfit prompt clauses. But the gate is shared prior art (DSPy relied on held-out validation in 2023), not the differentiator. What Regimes adds is upstream of the gate and downstream of it: the substrate makes the whole loop a first-class workflow, so failures are logged, diagnoses replay deterministically, candidate patches scope to typed seams, and every promote-or-discard decision is itself an event in the agent's history rather than a row in an external table. The approaches also differ in what they edit and where they point: GRASP gates a library of natural-language skills as the deployed artifact, in structured procedural environments (FHIR tasks, ALFWorld, WebShop); Regimes gates executable patches at typed pipeline seams on an event-sourced substrate, targets evidence reconciliation in long-context memory (the assemble-internal class, which has no direct analog in GRASP's setting), and treats the promoted prompt as a probe whose endpoint is a distilled guarded operator rather than as the destination. We do not compete on effect magnitude: GRASP reports 21 to 48 points in an unsaturated regime, whereas Regimes operates against a high-retrieval-accuracy regime (CONFIRM baselines 0.71 to 0.78, post-transform up to 0.88) where gains of that scale are mathematically excluded, a few held-out points is meaningful movement, and the failure analysis is the contribution. Read as an algorithm paper, Regimes overlaps heavily with GRASP and has weaker results; read as a systems paper, the overlap is one shared component and the contributions are largely disjoint. The latter reading is the intended one.

Held-out prompt and program optimization. DSPy \citep{khattab2023dspy} treats the natural-language portion of an LLM pipeline as parameters optimized against a development set, using held-out validation to compare candidates. Regimes' held-out-gated transform discovery is in this lineage. The difference is granularity and target: Regimes optimizes a typed transform at a specific seam selected by failure diagnosis, with an explicit regression-bounded acceptance rule, rather than a general pipeline prompt.

Failure taxonomies. MAST \citep{cemri2025mast} provides the first empirically grounded multi-agent failure taxonomy, with fourteen modes across three categories, validated by expert annotators and an LLM-as-judge pipeline. AgentDebug \citep{zhu2025agentdebug} introduces a modular agent-failure taxonomy and a debugging framework that traces failures to root causes. Regimes differs in two ways. Its taxonomy is single-agent, and more importantly it is executable: each regime maps to a specific action seam, so the taxonomy does not merely label failures but routes the loop's next edit.

The benchmark and the wall. LongMemEval \citep[ICLR 2025]{wu2024longmemeval} is the benchmark used here. It comprises 500 questions over scalable chat histories and tests five memory abilities, namely information extraction, multi-session reasoning, temporal reasoning, knowledge updates, and abstention. Abstention being a named ability there is why the abstention analysis in Section 5.6 is a recognized axis rather than an ad-hoc check. LongMemEval-V2 \citep{wu2026longmemeval2} extends the family to web-agent environment experience, which is a different axis from V1's chat-history reconciliation. We cite V2 to mark the family's continued activity, not as the setting of our result.

Substrate. Regimes is built on ActiveGraph \citep[``The Log is the Agent'']{nakajima2026log}, an event-sourced runtime whose graph state is a deterministic projection of an append-only event log, with a determinism contract and a content-addressed cache that records model and tool responses so replay performs no new model calls. That substrate paper discusses self-improving agents as an affordance the architecture enables, not a result it demonstrates. Regimes is the empirical follow-up that delivers, under held-out validation, what the substrate paper left as future work.

\section{The Regimes platform}

\subsection{Three taxonomies (terminology)}

We distinguish three categorizations, conflated at one's peril:

\begin{itemize}
\item Question type: the LongMemEval category. temporal-reasoning, knowledge-update, multi-session, single-session-preference, single-session-user, single-session-assistant.
\item Failure regime: the diagnosed reason a question failed. budget-truncation (evidence dropped at the context budget), assembly-crowding, assemble-internal (evidence present, answer wrong, the reconciliation wall), retrieval-signal-gap (the right turns never scored high enough), scoring-error.
\item Action seam: what Regimes may modify. A score-transform (re-weight retrieved-turn scores), an assembly-transform (reorder or filter selected turns), or a reader-prompt-transform (edit the reader's prompt fragments). All three are executable patch objects, differing in which pipeline stage they patch.
\end{itemize}

\subsection{The regime-to-seam mapping (Figure 1)}

The mapping is the organizing heuristic of the design: it selects which seam to attempt for a given failure, rather than searching all seams blindly. budget-truncation and assembly-crowding route to score- or assembly-transforms. assemble-internal routes to a reader-prompt-transform. retrieval-signal-gap and scoring-error have no seam that reaches them and are treated as true walls. The reader-prompt transform is the only seam that can touch the reconciliation wall, because it is the only one that changes how the reader uses evidence rather than which evidence it sees. We are careful about what the experiments support here: Section 5.8 reports evidence consistent with this mapping (the reader-prompt seam helped in every split, where assemble-internal was the dominant regime, and the score-transform seam generalized only in the single split where its target regime, budget-truncation, reached co-dominant share), but that evidence is indirect. The marginal contribution of the routing step itself, over a baseline that simply hands an author the failed examples and the held-out gate without regime labels or seam constraints, is not isolated by any experiment here; it is the primary open question, and the no-routing ablation that would settle it is stated as proposed work (Threat 11, Section 8.1).

\begin{figure}[!t]\centering
\includegraphics[width=\linewidth]{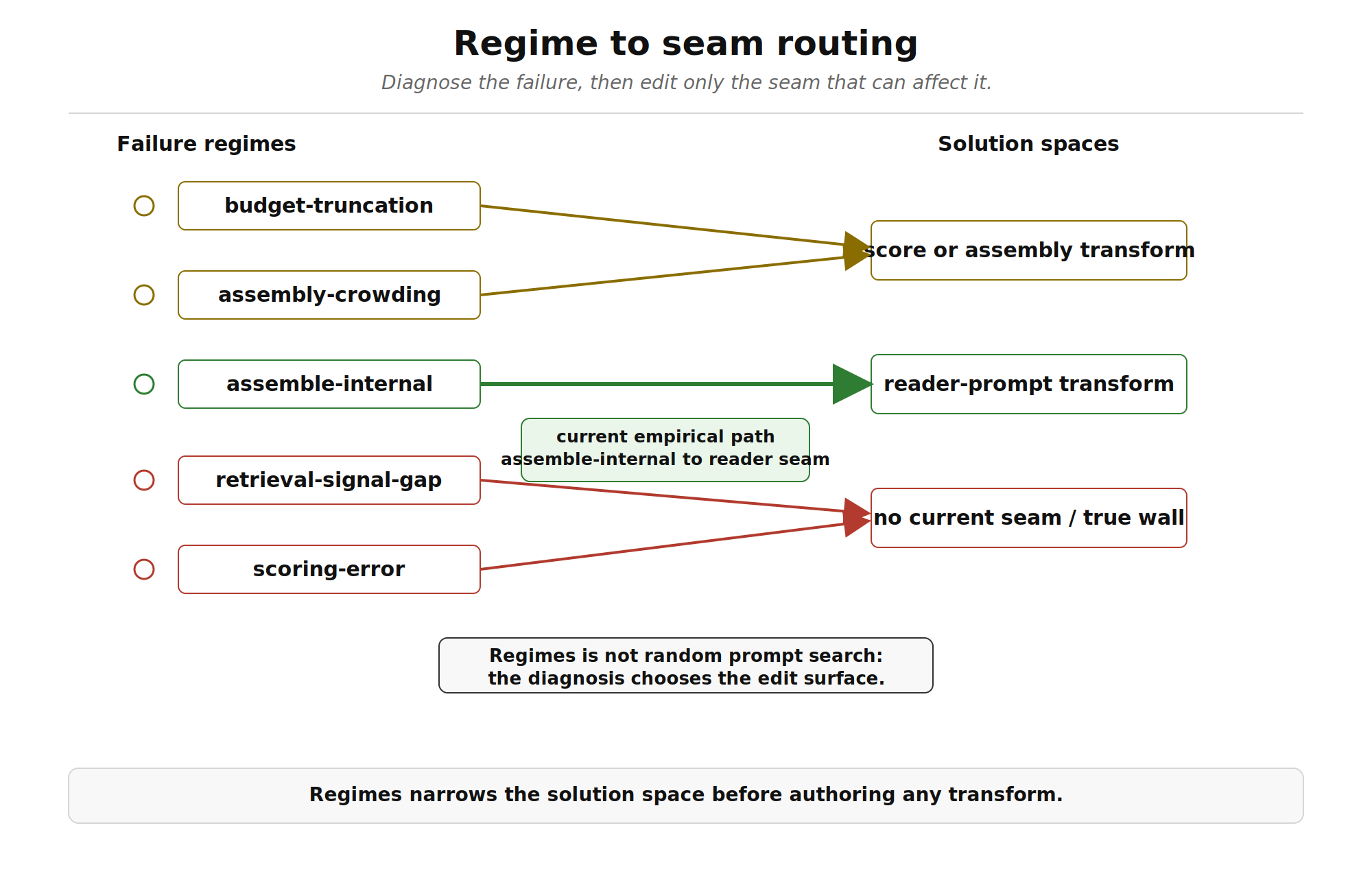}
\caption{Regime-to-seam routing. Left, the failure regimes. Right, the action seams. Arrows show the mapping: budget-truncation and assembly-crowding route to score- or assembly-transforms, assemble-internal routes to a reader-prompt-transform, and retrieval-signal-gap and scoring-error route to a "no seam, true wall" node.}
\label{fig:routing}
\end{figure}

\subsection{The loop (Figure 2)}

Each iteration runs four phases. Diagnose: run the agent on the OPTIMIZE split, classify each failure into a regime via fixed detectors (Section 6 reports the detection mechanism), build a regime histogram, and select the dominant optimizable, seam-reachable regime. Author: an LLM author drafts an executable transform of the type the selected regime routes to, given the relevant failure signals, which for reader-prompt transforms are the questions where evidence was present but the answer was wrong. Gate: static analysis (import whitelist, signature check, structural invariants, where a reader-prompt transform may edit values but not fabricate keys and may inject at most 2000 characters), then sandbox execution, then in-sample eval-diff on OPTIMIZE. Validate: run the candidate on the held-out CONFIRM split and promote only if CONFIRM does not regress. On promotion or exhaustion of a regime's attempts, rotate to the next seam-reachable regime, and stop only when all are exhausted, with a global iteration cap as backstop.

\begin{figure}[t]\centering
\includegraphics[width=\linewidth]{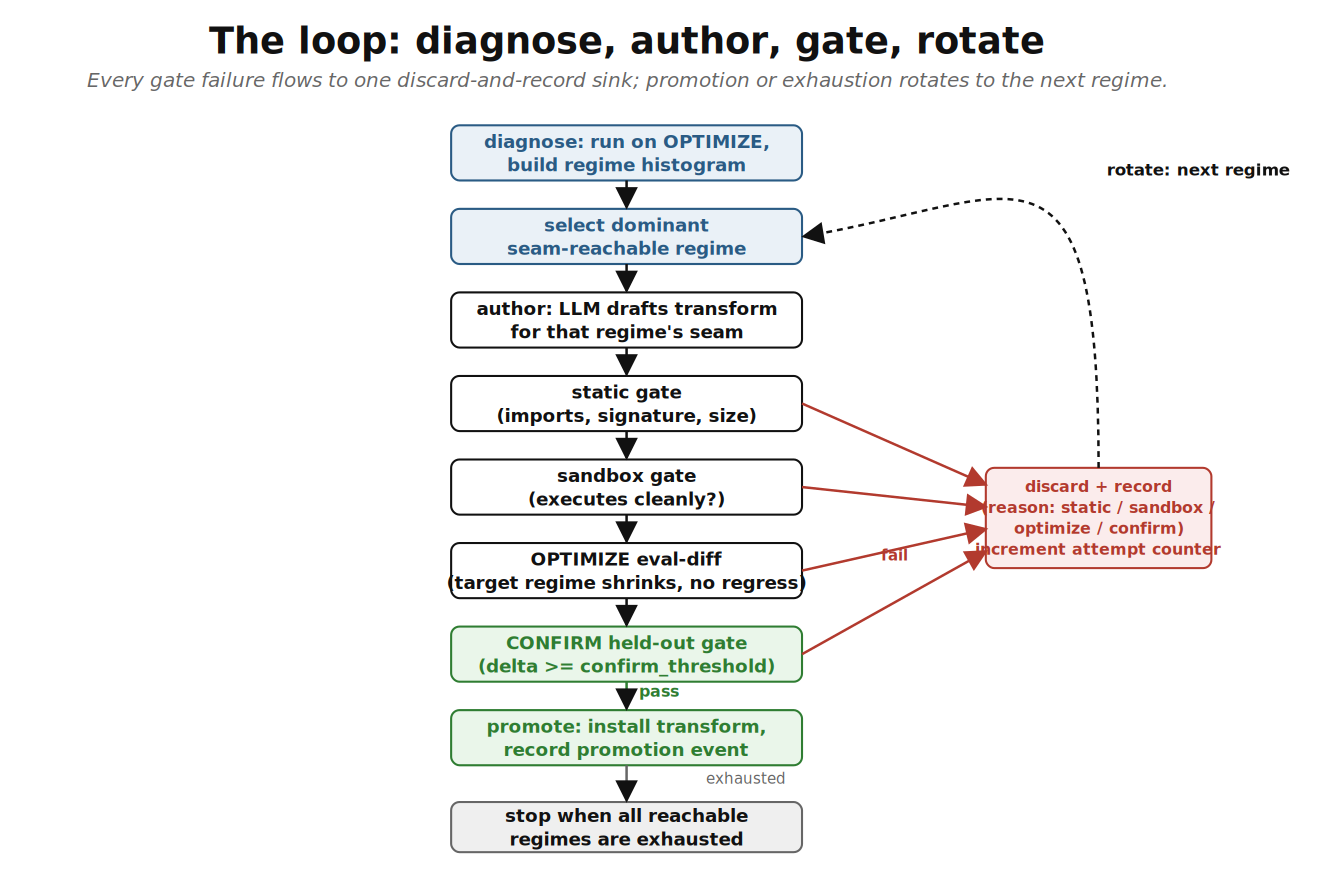}
\caption{Loop control flow with gates. The loop diagnoses failures into a regime histogram, selects the dominant seam-reachable regime, has the author draft a candidate, and passes it through the static, sandbox, in-sample OPTIMIZE, and held-out CONFIRM gates in order. Any gate failure routes to a single discard-and-record sink that increments the regime's attempt counter identically; on promotion or attempt exhaustion the loop rotates to the next reachable regime, stopping only when all are exhausted.}
\label{fig:loop}
\end{figure}

Pseudocode:

\begin{verbatim}
for regime in reachable_regimes_by_priority(histogram):
    for attempt in range(max_attempts):
        candidate = author(regime, failure_signals[regime], current_state)
        if static_check(candidate) fails:   record(discard, "static"); continue
        if sandbox_run(candidate)  fails:   record(discard, "sandbox"); continue
        opt_delta = eval_optimize(candidate)
        if not (target_regime_shrinks(opt_delta) and not overall_regresses(opt_delta)):
            record(discard, "optimize"); continue
        confirm_delta = eval_confirm(candidate)     # vs current pipeline minus candidate; binding gate
        if confirm_delta >= confirm_threshold:       # confirm_threshold default 0.0
            promote(candidate); record(promote, confirm_delta); break
        else:
            record(discard, "confirm_regression")
    # exhausted this regime's attempts -> rotate
stop_reason = "all_reachable_regimes_exhausted"     # never None (runner backstop)
\end{verbatim}

The \texttt{confirm\_delta >= confirm\_threshold} rule is the promotion gate; the reported runs used the default \texttt{confirm\_threshold = 0.0}. Section 5.7 shows this default is too permissive at high baselines, where it admits within-noise promotions, and Section 8.2 builds the fix on top of this existing threshold hook.

\subsection{The target interface}

The loop is decoupled from any specific agent behind a Target interface: Target = EvalBackend + ActionSpace + RegimeTaxonomy + outcome\_summary. The loop, gates, rotation, and held-out discipline operate only against this interface. A target supplies its own evaluator, action seams, failure detectors, and outcome summarizer.

\subsection{Binding gates, installed before widening the action space}

The held-out gate is binding. A transform that improves in-sample but regresses, or fails to clear threshold, on CONFIRM is automatically discarded with the reason recorded. The gate computes confirm\_delta by measuring CONFIRM accuracy with the candidate installed, then reverting only that candidate (a keyed removal, leaving any previously promoted transforms in place) and measuring again, so confirm\_delta is a marginal over the current deployed state: the candidate must not regress the pipeline that already carries the earlier promotes. These per-candidate marginals do not telescope to the final deployed accuracy, for two reasons. First, each is a fresh pair of held-out evaluations subject to reader nondeterminism (Section 5.1), not an arithmetic increment on a running total. Second, later reader-prompt transforms re-work the same assemble-internal questions earlier ones already addressed rather than adding disjoint gains (Section 5.7), so a positive marginal at step $k$ can be measured against a confirm\_before that already fixed many of the same questions. The consequence is that the gate prevents a candidate from regressing the deployed state at its own promotion step but does not guarantee monotonic improvement of the cumulative deployed state across a long promotion sequence: a run of individually non-regressing, within-noise promotes can leave the final state no higher than an intermediate one. That gap is exactly the over-promotion failure in Section 5.7 and the motivation for the plateau rule in Section 8.2. This guardrail was installed before the action space was widened to include reader-prompt transforms. The reason is deliberate: a wider action space, especially free-text prompt edits, overfits far more easily than score re-weighting, and a wide seam with a non-binding held-out check would promote overfit transforms and report false wins. Held-out usage: CONFIRM is consulted once per candidate that clears the in-sample gate. The per-split count of candidates reaching CONFIRM is reported in the candidate-funnel table (Table 1c) as the multiplicity disclosure (Section 5.5).

\subsection{Chaotic-mock hardening (robustness of the loop itself)}

A real LLM author produces malformed code, catastrophic regressions, and mixed sequences of discards, static rejections, and sandbox crashes. The loop's control flow must rotate and terminate cleanly under that distribution. We test it with a chaotic mock author that emits the full failure distribution: valid-but-discarded, valid-but-regressing, syntactically invalid, and sandbox-crashing. All candidate outcomes flow through a single failure sink that increments the regime's attempt counter identically, and a runner backstop makes a null termination structurally impossible. Section 4.4 reports how this discipline was forced on us by real runs.

\section{Generalization test: a second target}

\subsection{Extraction under a byte-identical constraint}

Regimes began as a LongMemEval-specific script. We extracted the Target interface and re-homed LongMemEval behind it under a hard constraint: byte-identical loop behavior before and after. Here byte-identical means the serialized event log of a run, which is the ordered sequence of events the loop emits (diagnose, author, gate outcomes, promote or discard, rotate) with their recorded fields, is byte-for-byte equal before and after the refactor on the same inputs and seed. It does not mean model outputs are regenerated identically across API calls. It means the loop's own emitted trace is identical, which is the property that certifies the extraction preserved control-flow semantics rather than quietly changing them. This is the event-log determinism of the ActiveGraph substrate \citep{nakajima2026log}. Section 6 states the narrower determinism claim around LLM nondeterminism.

\subsection{The second target as a unit test for the abstraction}

We built a second target: a text-to-SQL agent with a different action space (edit the prompt, not re-rank turns), different failure modes (wrong join, missing GROUP BY), and a different evaluator (execute SQL, compare result rows). It runs through the unchanged loop machinery and produces the same event sequence.

Building it surfaced four couplings where the generic loop still silently assumed LongMemEval's shape, including a load-bearing one where the promotion gate computed whether the target failure shrank using LongMemEval's taxonomy, which would have broken every future target. None were visible with a single target. The fix was deletion: after re-homing the couplings, the SQL target's reimplemented eval-diff and sandbox gate collapsed to eight-line delegations. Deletion is the proof of absorption. If the special-case code does not shrink to a call after generalizing, nothing was generalized.

\subsection{Three senses of generality}

\begin{itemize}
\item Interface generality, shown: the loop operates over multiple Target implementations.
\item Control-flow generality, shown: gates, rotation, and candidate handling behave identically across targets, evidenced by byte-identical event logs (Section 4.1).
\item Empirical-improvement generality, not claimed: that the loop improves multiple tasks under held-out validation. The SQL target was not run to a held-out improvement, because a frontier reader scored near ceiling on the available SQL fixture, leaving no seam-reachable headroom. We make no cross-task empirical-improvement claim. The empirical result in Section 5 is LongMemEval-specific.
\end{itemize}

\subsection{Four integration gaps, each found by a real run, each missed by a green test suite}

After widening the action space and passing every mock test, the new reader-prompt seam would not fire on real runs. Across several costly runs the loop kept drafting only score-transforms and stopping. Each run revealed one gap.

\begin{enumerate}
\item Stale taxonomy flag. assemble-internal was still marked not-seam-reachable, so the loop never routed to it. Fix: derive seam-reachability from the regime-to-seam map as a single source of truth.
\item Global stop before rotation. The loop exhausted the dominant regime and stopped the entire loop, never rotating. Fix: exhaust then rotate, and stop only when all reachable regimes are exhausted.
\item Seam not wired through evaluation. Reader-prompt transforms were installed into a pipeline the eval path never read, so they changed nothing, scored zero, and were discarded every time. This would have produced a convincing false negative, that the loop cannot cross the wall, when the true cause was that the transform never reached the reader.
\item A malformed candidate killed the loop. A static rejection fell out of the iteration without rotating or stopping cleanly, leaving a null stop reason.
\end{enumerate}

The general lesson: the mock author produced only clean candidates, so a passing suite said nothing about robustness to a real author's chaos. The fix was the chaotic-mock hardening of Section 3.6. The decisive test reproduced the exact real failure: discard, discard, static-reject, then it must rotate and must not null-terminate. Build fixtures from real artifacts and make fakes at least as messy as production. A chaotic mock would have caught all four gaps at once, for free.

\section{Application to LongMemEval and results}

\subsection{Setup}

Two scoping notes belong up front, because they bound how the results in this section should be read. First, regime labels are offline diagnostic labels, not deployment-time signals: the regime classifier reads gold evidence locations to decide which repair seam to test during benchmark improvement (full mechanism in Section 6), so any claim that depends on those labels (the localization in Section 5.6, the seam-share analysis in Section 5.8) holds "under this classifier" and is only as strong as the label quality, whose human validation is the open item in Table 2. Second, on noise: cached replay buys exact reproducibility, not a noise estimate. Characterizing the reader noise band requires fresh API calls, which is the expensive thing the cache does not do. We estimate the band at roughly plus or minus 0.02 to 0.04 from cross-run baseline variation rather than from repeated same-split evaluation, and we flag where that band is close to a reported effect (seed 23's +0.05).

Each run uses a stratified split of the 500-question LongMemEval-S set into OPTIMIZE = 50, on which the loop diagnoses, drafts, and runs in-sample eval-diffs, and CONFIRM = 100, held out and used only to validate a candidate that already passed in-sample gates. The fresh-split runs use disjoint OPTIMIZE and CONFIRM samples drawn by a seeded stratified split. CONFIRM does not overlap OPTIMIZE within a run. These are independent loop runs over fresh stratified splits, but not independent samples from an unlimited population, since all splits are drawn from the same 500-question benchmark pool.

A third note concerns the regime histogram that drives routing. It is computed from the failures on the 50-question OPTIMIZE split, which number 6 to 21 across the five splits, so the dominant-regime ranking carries genuine small-sample variance. In four of five splits assemble-internal dominates by a clear margin (9 of 10 failures in seed 7, 8 of 13 in seed 11, 11 of 14 in seed 23, 13 of 21 in seed 5), so routing to the reader-prompt seam is robust there. Seed 101 is the exception: on the smallest pool (6 failures) assemble-internal and budget-truncation tie at 3 each, so its routing is genuinely noise-sensitive. That tie is consistent with seed 101 being the noisiest split on every other axis (highest baseline, near-null final state, the over-promotion case), and it is the same split where budget-truncation reached enough share for a score-transform to be promoted (Section 5.8). We do not quantify routing stability beyond reporting these pool sizes; the five-split replication provides indirect robustness rather than a stability guarantee (Threat 9).

Reader: claude-sonnet-4-6, decoded greedily at temperature 0.0 with max\_tokens 1024, tool-free. The LLM author uses the same model, claude-sonnet-4-6, at temperature 0.2 with max\_tokens 2048. Embeddings use OpenAI text-embedding-3-small (a deterministic hash-embedder fallback exists for offline replay). Scoring runs on the deterministic ActiveGraph substrate. A single baseline evaluation was run per split. The residual reader variance noted above is hosted-model nondeterminism rather than sampling temperature: even at temperature 0.0, request batching, hardware, and silent endpoint updates can change greedy outputs (Threat 2; Section 6 gives the narrowed determinism claim).

Baseline reporting. "Baseline" can mean the OPTIMIZE-split or the CONFIRM-split baseline, and they differ, so Table 1 and Table 1b separate them. None of these are the roughly 0.86 full-500 headline from prior LongMemEval work. They are 50/100 split baselines, and every result is paired before-and-after movement on a split, not a new headline score.

Promotion thresholds. The in-sample gate requires the candidate to shrink the target regime and not regress overall on OPTIMIZE. The held-out gate compares the CONFIRM delta against a configurable \texttt{confirm\_threshold} whose default is 0.0, so by default a candidate is promoted when it does not regress held-out. The code exposes this threshold as a tunable, with an inline note that callers may set it to roughly +0.02 to require clearing the reader noise band; the runs reported here used the default of 0.0. Section 5.7 shows that the default 0.0 is too permissive at high baselines, where it admits within-noise promotions, and Section 8.2 builds the fix on top of this existing threshold hook.

\subsection{Headline result}

The strongest single split is seed 5, where the promoted reader-prompt repair moved held-out accuracy from 0.78 to 0.88 (+0.10), with 11 wrong-to-right and 1 right-to-wrong flips (McNemar exact two-sided p = 0.006). The original fresh split, seed 7, moved 0.74 to 0.82 (+0.08), 10 wrong-to-right and 2 right-to-wrong (p = 0.039). These two splits are individually significant under a simple paired test (Section 5.5 notes that seed 5's value is not adjusted for its four-promote sequential structure). The fuller result is the five-split replication in Section 5.3, where the effect is consistent in direction while several individual splits are underpowered.

\subsection{Multi-seed replication (Table 1b, Figure 3)}

We ran five fresh stratified splits (seeds 7, 11, 23, 5, 101; seed 7 is the original Run 3 split, the others are arbitrary non-sequential values chosen so the replication does not depend on seed 7's initial motivating run) and report all five, including the split that ended near zero. For splits where the loop promoted more than one transform in sequence, we report the final state of the run, which is the state a deployment would inherit, and we record the intermediate promotes in the notes rather than select the best one. All five fresh splits retain final paired correctness vectors. Seed 7 additionally has the richer per-question metadata needed for the per-type, abstention, and regime-localization tables in Section 5.6.

Table 1b. Five fresh stratified splits, final state. Columns are ordered to foreground the flip asymmetry (delta, w->r, r->w), which is the most consistent signal; baselines and the post-hoc McNemar p follow.

\begin{center}\small
\resizebox{\textwidth}{!}{%
\begin{tabular}{llllllllll}
\toprule
split & delta & w->r & r->w & discordant & CONFIRM base & CONFIRM post & OPT base & promotes (total) & post-hoc McNemar p \\
\midrule
seed 7 (Run 3) & +0.08 & 10 & 2 & 12 & 0.74 & 0.82 & 0.80 & 1 & 0.039 * \\
seed 11 & +0.06 & 8 & 2 & 10 & 0.77 & 0.83 & 0.74 & 2 & 0.109 \\
seed 23 & +0.05 & 7 & 2 & 9 & 0.71 & 0.76 & 0.72 & 1 & 0.180 \\
seed 5 & +0.10 & 11 & 1 & 12 & 0.78 & 0.88 & 0.58 & 4 & 0.006 * \\
seed 101 & +0.01 & 7 & 6 & 13 & 0.78 & 0.79 & 0.88 & 6 & 1.000 \\
\bottomrule
\end{tabular}}
\end{center}

\begin{itemize}
\item individually significant at alpha = 0.05. "Promotes (total)" counts all accepted candidate promotions in the run; in a multi-promote split the final deployed state is the cumulative stack of those promotions, and the reported delta and flips are for that final state. Seed 101's 6 total promotions are 5 reader-prompt transforms (the sequence in Section 5.7) plus 1 budget-truncation score-transform (Section 5.8); Table 1c reports the same total of 6. Seed 5's OPTIMIZE baseline of 0.58 is low, but its CONFIRM (held-out) baseline of 0.78 is in the normal band, so its +0.10 is not a low-baseline artifact. We flag it directly: seed 5 drives the only p < 0.01 and the most promotes, and it also has the widest OPTIMIZE-to-CONFIRM baseline gap (20 points, where other splits are within a few points). The most likely explanation is 50-question OPTIMIZE-split variance rather than anything systematic, since the held-out CONFIRM behavior is in-band and the four promotes climb cleanly (Section 5.7); a skeptical reader should nonetheless weight the pooled direction across splits over this single strongest split.
\end{itemize}

As a descriptive pooled summary across all five final-state splits, the discordant counts are 43 wrong-to-right versus 13 right-to-wrong (McNemar exact two-sided p = 7.3e-05). As a sensitivity check, excluding seed 101 (the over-promotion split, Section 5.7) gives 36 versus 7 and p = 9.0e-06. The per-seed rows are the primary evidence; the pooled discordances are a descriptive summary only, carrying the same-pool caveat from Section 5.1, and we do not treat the pooled p as a population-level significance result. We report the all-five pooled value rather than the seed-101-excluded one as the headline descriptive number, because excluding the weakest split would inflate even the descriptive summary.

Reading the table: four of five splits are clearly positive, from +0.05 to +0.10. The fifth, seed 101, ended at +0.01, which is within noise. Two of five splits are individually significant. Seeds 11 and 23 are individually underpowered, with twelve or fewer discordant pairs each, where clearing p < 0.05 is not achievable at counts like 7 versus 2. Seed 101 is a different case: it has thirteen discordant pairs, more than seed 5 or seed 7, so it is not underpowered. Its non-significance is a genuine null, with near-symmetric flips (7 wrong-to-right versus 6 right-to-wrong, p = 1.0 by exact McNemar), arising from the over-promotion mechanism in Section 5.7 rather than from insufficient sample size. The pattern is a consistent positive direction across all splits; the pooled discordance test is a descriptive summary rather than independent-sample evidence, because the splits share the same benchmark pool (Section 5.5). The regression count is the other stable signal: in the four clean splits, right-to-wrong flips were 2, 2, 2, and 1. The transforms reliably break almost nothing while fixing seven to eleven each. Seed 101's drift to six right-to-wrong flips is the over-promotion signal analyzed in Section 5.7.

Three tiers of replication, to avoid over-reading the word replicated:

\begin{itemize}
\item Transform-discovery repeatability (the two earlier fixed-split runs, Section 5.4): two runs on the same held-out questions independently authored reconciliation transforms that both cleared that set. This is evidence the discovery is repeatable, not independent evidence of benchmark-level generalization, since the split is shared.
\item Held-out-sample robustness (the five fresh splits here): different stratified splits, independently authored transforms, positive in every split.
\item Pooled discordance summary across all five reported splits: the descriptive pooled count, carrying the same-pool caveat.
\end{itemize}

\begin{figure}[t]\centering
\includegraphics[width=\linewidth]{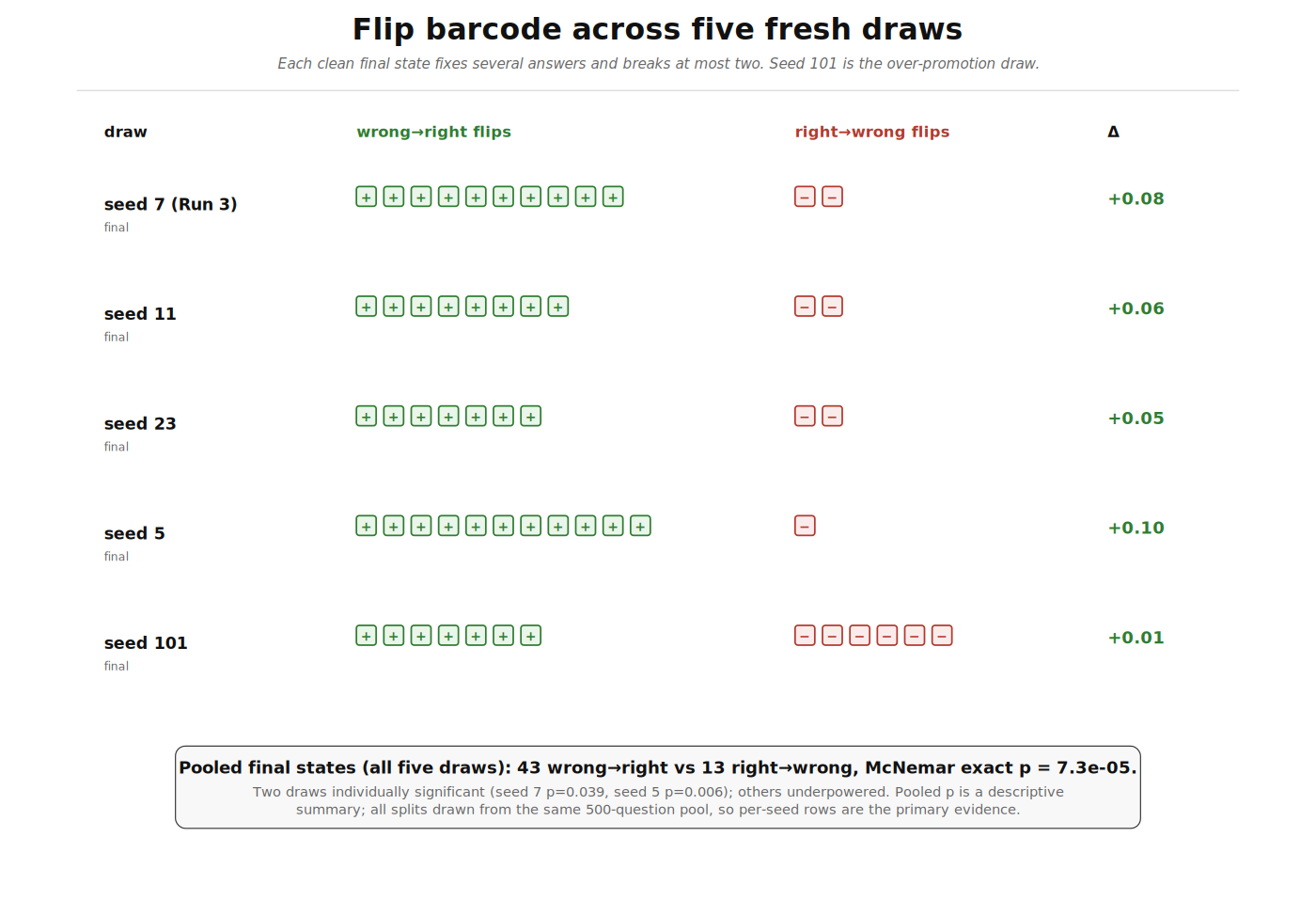}
\caption{Flip barcode across the five fresh stratified splits. Each row shows one split's final-state flips: green squares are wrong-to-right, red squares are right-to-wrong, with the per-split held-out delta at right. Pooled across splits, 43 wrong-to-right versus 13 right-to-wrong; this pooled count is a descriptive same-pool summary (Section 5.5). Seed 101's six right-to-wrong flips are the over-promotion signal (Section 5.7).}
\label{fig:barcode}
\end{figure}

\subsection{The two earlier fixed-split runs (Table 1)}

Before the fresh-split replication, two runs used a single fixed split. They predate the per-question persistence fix (Section 6), so only their aggregate CONFIRM delta survives.

Table 1. Earlier fixed-split runs.

\begin{center}\small
\resizebox{\textwidth}{!}{%
\begin{tabular}{lllllll}
\toprule
run & split & OPT base & OPT post & CONFIRM delta & flips & significance \\
\midrule
1 & fixed & 0.70 & 0.88 & +0.04 & not retained & not retained \\
2 & fixed (same held-out as 1) & 0.78 & 0.86 & +0.03 & not retained & not retained \\
\bottomrule
\end{tabular}}
\end{center}

These reproduce the direction of the effect with independently authored transforms on a shared held-out set: directional support, not independent significance. In-sample gains across runs disagree widely (one reached in-sample +0.18) while held-out deltas stay in a tight positive band, the signature of the gate discounting overfit (Section 6).

\subsection{Significance, interval, and selection-pressure disclosure}

For seed 7, a paired nonparametric bootstrap over the 100 held-out questions (percentile method, 10,000 resamples) gives mean paired difference +0.08, 95\% CI [+0.02, +0.15].

Statistical disclosures. McNemar's exact test was not pre-registered; it is the standard choice for paired binary outcomes and is reported as a post-hoc characterization. The CI is a percentile bootstrap over questions; with n = 100 and twelve discordant pairs it should be read cautiously. The pooled five-split summary treats all roughly 500 held-out questions as paired pairs, but the five splits come from the same 500-question pool, so the held-out sets overlap in membership across seeds. The pooled McNemar treats pairs as independent, which overstates effective n. We therefore present the per-seed rows as the primary evidence, with the consistent positive direction across five splits as the main claim and the pooled p as a descriptive summary carrying this caveat. The five per-split tests are also a family: under a Bonferroni correction for five tests (adjusted alpha 0.01) seed 5 (p = 0.006) remains significant and seed 7 (p = 0.039) does not, so the family-wise-supported claim is one significant split, not two. Because the tests were not pre-registered with a primary endpoint, both the per-split and family-level significance should be read as exploratory characterization of an effect whose main evidence is its directional consistency, not as a confirmatory significance result.

Selection pressure and multiplicity (Table 1c). Because the held-out gate compares candidates, a single McNemar p understates effective multiplicity if many candidates were scored against CONFIRM. We disclose the candidate funnel per split, taken from the run logs and reports.

Table 1c. Candidate funnel per split.

\begin{center}\small
\resizebox{\textwidth}{!}{%
\begin{tabular}{llllll}
\toprule
split & authored & static-rejected & discarded on OPTIMIZE & discarded on CONFIRM & promoted \\
\midrule
seed 7 & 4 & 0 & 3 & 0 & 1 \\
seed 11 & 9 & 1 & 5 & 1 & 2 \\
seed 23 & 7 & 0 & 6 & 0 & 1 \\
seed 5 & 10 & 0 & 6 & 0 & 4 \\
seed 101 & 14 & 0 & 7 & 1 & 6 \\
total & 44 & 1 & 27 & 2 & 14 \\
\bottomrule
\end{tabular}}
\end{center}

The dominant filter is the OPTIMIZE gate, not CONFIRM. Across the splits, almost all discards happened on the in-sample gate (target regime did not shrink, or multi-session or overall regressed), with only two CONFIRM regressions total (one in seed 11 and one in seed 101, both budget-truncation score-transforms). Few candidates reach the held-out check and clear it only to be rejected, so CONFIRM selection pressure is low, which bounds the multiplicity concern. Seed 7, the most analyzed split, authored only four candidates and discarded three of them on OPTIMIZE before a single promote, which is the low-pressure case in its clearest form. Seed 101 is the exception in promote count, not in CONFIRM pressure, and is analyzed next.

One form of multiplicity is not bounded by the funnel and should be stated directly. In the multi-promote splits (seed 5 with four promotes, seed 101 with six), the same 100 CONFIRM questions are consulted once per promotion decision. The marginal-contribution gate reduces but does not eliminate adaptive data use across that chain: the author at step k works in a state where steps 1 to k-1 were already CONFIRM-compatible, so the same held-out set is being used across several dependent selection decisions, which can inflate Type I error even when each individual step is non-retentive of labels. Seed 5's p = 0.006 is computed on the endpoint of a four-step adaptive process and is reported without adjustment for that structure. We therefore frame the two-split significance result conservatively: two splits are individually significant under a simple paired test that does not adjust for sequential promotion, and a clean test would re-evaluate a promoted pipeline on a third split withheld from both OPTIMIZE and CONFIRM (proposed in Section 8.1).

\begin{figure}[t]\centering
\includegraphics[width=\linewidth]{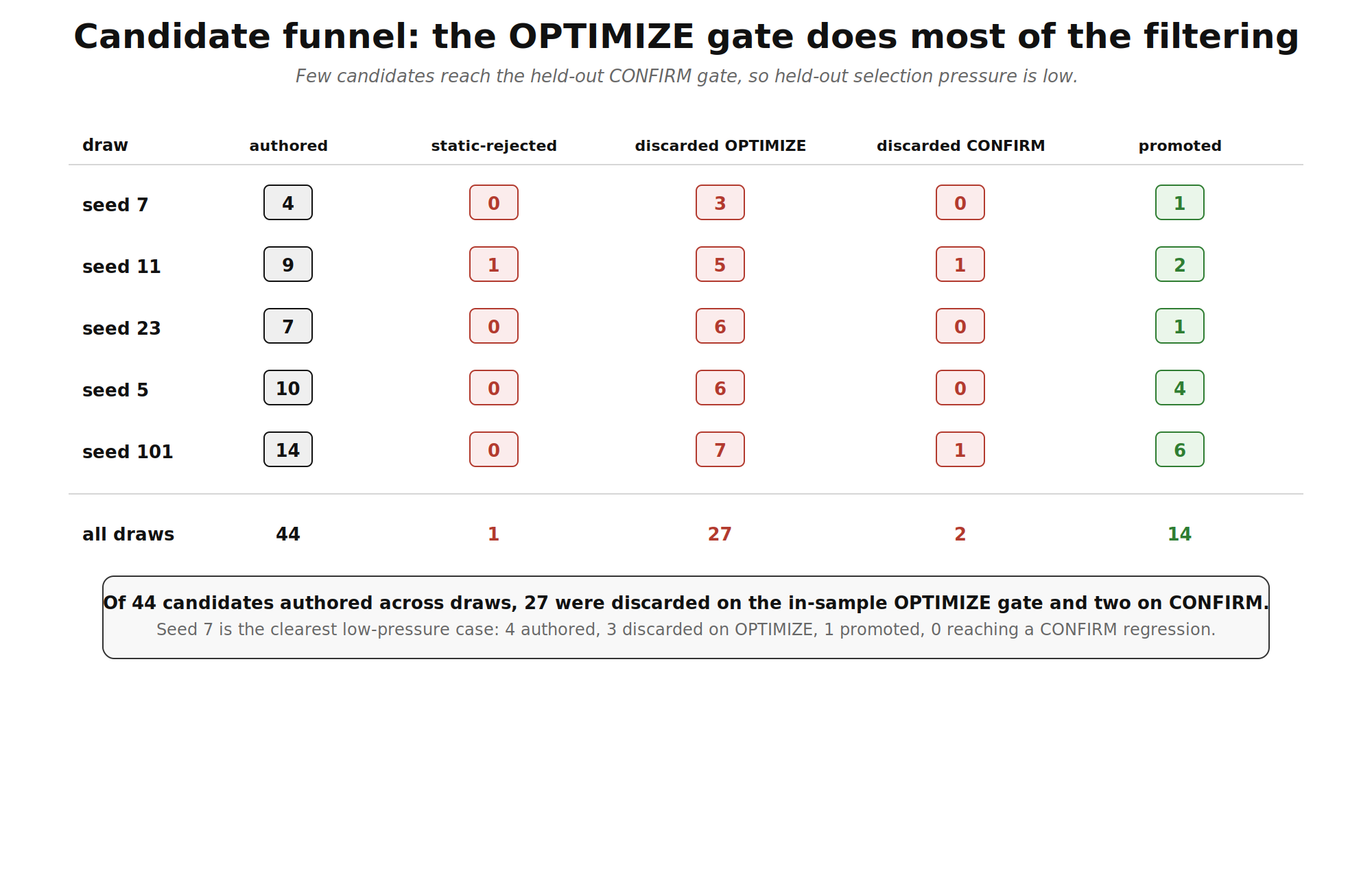}
\caption{Candidate funnel per split: authored, static-rejected, discarded on OPTIMIZE, discarded on CONFIRM, and promoted. Across splits, almost all discards happen on the in-sample OPTIMIZE gate, with two CONFIRM regressions total, so held-out selection pressure is low. Seed 7 is the clearest small-volume case (4 authored, 3 discarded on OPTIMIZE, 1 promoted).}
\label{fig:funnel}
\end{figure}

Reader nondeterminism, formally. The roughly plus or minus 0.02 to 0.04 band (Section 5.1) is the same order as the two fixed-split effects, so the fixed-split +0.03 and +0.04 are reported as directional only; the per-seed fresh-split effects of +0.05 to +0.10 are what carry the claim, with the pooled discordance count a descriptive summary only (Section 5.5).

\subsection{The four held-out tables (seed 7)}

Seed 7 is the split with fully persisted per-question CONFIRM outcomes, including regime labels and abstention flags, so the detailed tables are computed on it. All five splits retain final paired correctness vectors, but only seed 7 has the richer per-question metadata needed for the tables below.

Per-type held-out delta (n = 100):

\begin{center}\small
\begin{tabular}{llll}
\toprule
question type & baseline & post & delta \\
\midrule
multi-session & 16 & 21 & +5 \\
single-session-preference & 1 & 3 & +2 \\
single-session-user & 13 & 14 & +1 \\
knowledge-update & 15 & 15 & 0 \\
single-session-assistant & 11 & 11 & 0 \\
temporal-reasoning & 18 & 18 & 0 \\
\bottomrule
\end{tabular}
\end{center}

We observe no per-category regression on this held-out split, though several categories are too small for strong conclusions, for example single-session-preference. The confirm-gate protects only overall accuracy, so a hidden per-type regression was possible, and none is observed. Gains concentrate in multi-session.

Abstention. The transform's central instruction, which is to trust the retrieved context and not hedge with "I cannot find," is exactly what could increase false positives when evidence is genuinely absent. On the held-out abstention questions (n = 6), baseline 6 correct, transform 6 correct, delta zero. We report no observed abstention cost. With n = 6 this is reassuring, not conclusive (Threat 3; Section 8 proposes a dedicated abstention stress set). Abstention being one of LongMemEval's five named abilities \citep{wu2024longmemeval} is why we treat it as a first-class axis.

Flip table (held-out): 10 wrong-to-right, 2 right-to-wrong, net +8, roughly five fixes per break.

Localization (net flips by baseline regime, by oracle-derived regime labels): assemble-internal +8 and minus 0, budget-truncation +2 and minus 0, correct +0 and minus 2. By these labels, gains localize to the target regime, with modest spillover, and the two regressions are baseline-correct questions. One caveat keeps this from being more surprising than it is: assemble-internal is defined partly as "evidence present, answer wrong," and a reader-prompt fix is exactly the intervention most able to move that bucket, so some of the localization is true by construction. The bucket the loop targets is the one structurally most amenable to a reader-side repair. Correctness being scored by the independent LongMemEval judge protects the numbers themselves; the point here is only that the mechanism is intuitive rather than unexpected. The budget-truncation spillover flips may reflect genuine cross-regime help or oracle-label noise (Threat 4 and the classifier human-validation, Table 2).

The paired movement behind these numbers, for seed 7: of the 100 held-out questions, 72 stay correct, 2 go correct to wrong, 10 go wrong to correct, and 16 stay wrong. The 10-versus-2 discordant split is the source of the flip counts and the McNemar test.

\subsection{Over-promotion: the seed-101 finding}

Seed 101 is the most informative split in the study, and we present it as a diagnostic result rather than only a weak split. It shows precisely how an autonomous loop fails when its stopping rule is too loose: the loop can keep promoting small, noisy gains unless the held-out threshold and a stopping criterion are stricter. That is a concrete, actionable finding about the loop itself, and it motivates the fix in Section 8.2. It had a high OPTIMIZE baseline of 0.88, which leaves little real headroom, and the loop promoted five transforms in sequence. The per-promote held-out deltas were +0.01 (7 wrong-to-right, 6 right-to-wrong), +0.07 (9, 2), +0.09 (13, 4), +0.00 (5, 5), and +0.01 (7, 6). The middle promotes were real improvements. The loop then kept promoting after the gains were exhausted, and the last two promotes were coin-flips, dragging the final state back to +0.01.

The contrast with seed 5 is the key to the diagnosis. Seed 5 also promoted multiple times, four in its case, but it climbed cleanly: per-promote held-out deltas of +0.08 (8 wrong-to-right, 0 right-to-wrong), +0.05 (6, 1), +0.08 (9, 1), and +0.10 (11, 1), with right-to-wrong flips staying at zero or one throughout and the final promote being the best. Seed 101 peaked at its third promote (+0.09) and then decayed, with right-to-wrong flips climbing back to six. The problem is not that a split promotes multiple times. The problem is promotes after the gain has plateaued.

A note on what these per-promote deltas measure, since they do not sum to the final state and should not be read as if they did. Each per-promote delta is the candidate's CONFIRM accuracy minus the deployed state with that candidate reverted (its earlier promotes still installed), so it is a marginal over the previous deployed state, not over the bare split baseline. The reason it nonetheless does not telescope is that the previous deployed state itself did not compound: because the earlier reader-prompt transforms re-worked the same assemble-internal questions, the pre-candidate deployed accuracy held near 0.79 across the sequence rather than ratcheting up as transforms were promoted. The marginals are therefore real per-step measurements against a deployed state that stayed roughly flat, not sequential increments on a rising total, and they do not telescope: seed 5's +0.08, +0.05, +0.08, +0.10 do not add to its +0.10 final state, and seed 101's values do not add to its +0.01. For the same reason the per-promote wrong-to-right and right-to-wrong counts are not disjoint draws from a single shrinking pool of baseline-wrong questions and should not be summed: each pair is measured between consecutive deployed states (state $k\!-\!1$ versus state $k$) on the full CONFIRM set, so a question the earlier transforms re-work can be counted again at a later step, and the cumulative count across promotes can exceed the split's baseline-wrong total without contradiction. The final deployed accuracy of each split is the CONFIRM accuracy of the pipeline state after the last promotion (0.88 for seed 5, 0.79 for seed 101), not the sum of the per-promote deltas. This is why the over-promotion in seed 101 is a story about non-compounding promotes rather than a stack of gains that later collapsed: the later reader-prompt transforms re-worked the same assemble-internal questions the earlier ones had already addressed, so they cleared the noise-level gate against a flat baseline without adding new held-out wins. The loop did not stop because gains plateaued; it stopped on no\_optimizable\_regime\_remaining, having exhausted the seam-reachable regimes in the histogram. A plateau-aware stopping rule (Section 8.2) would have halted it several promotes earlier, at the +0.09 peak.

The cause is a stopping-criterion weakness. The CONFIRM gate compares the held-out delta against a \texttt{confirm\_threshold} whose default is 0.0, so a transform that nudges accuracy by +0.01 on a noisy held-out set clears the gate and is promoted, even when its flip counts (7 versus 6) are indistinguishable from noise. With the threshold at its default and no rule to stop when held-out gains plateau, the loop over-promotes at high baselines where headroom is small. We report seed 101 at its final state (+0.01) rather than at its best promote (+0.09), because reporting the best would be selection on the outcome.

This is a genuine limitation and it has a clean fix, stated as a hypothesis in Section 8.2: a held-out-plateau stopping criterion that halts promotion when successive promotes stop producing flip-count improvements that exceed the noise band. The finding also connects to the prompt-cruft note in Section 9, since over-promotion accumulates prompt text without accumulating accuracy.

\begin{figure}[t]\centering
\includegraphics[width=\linewidth]{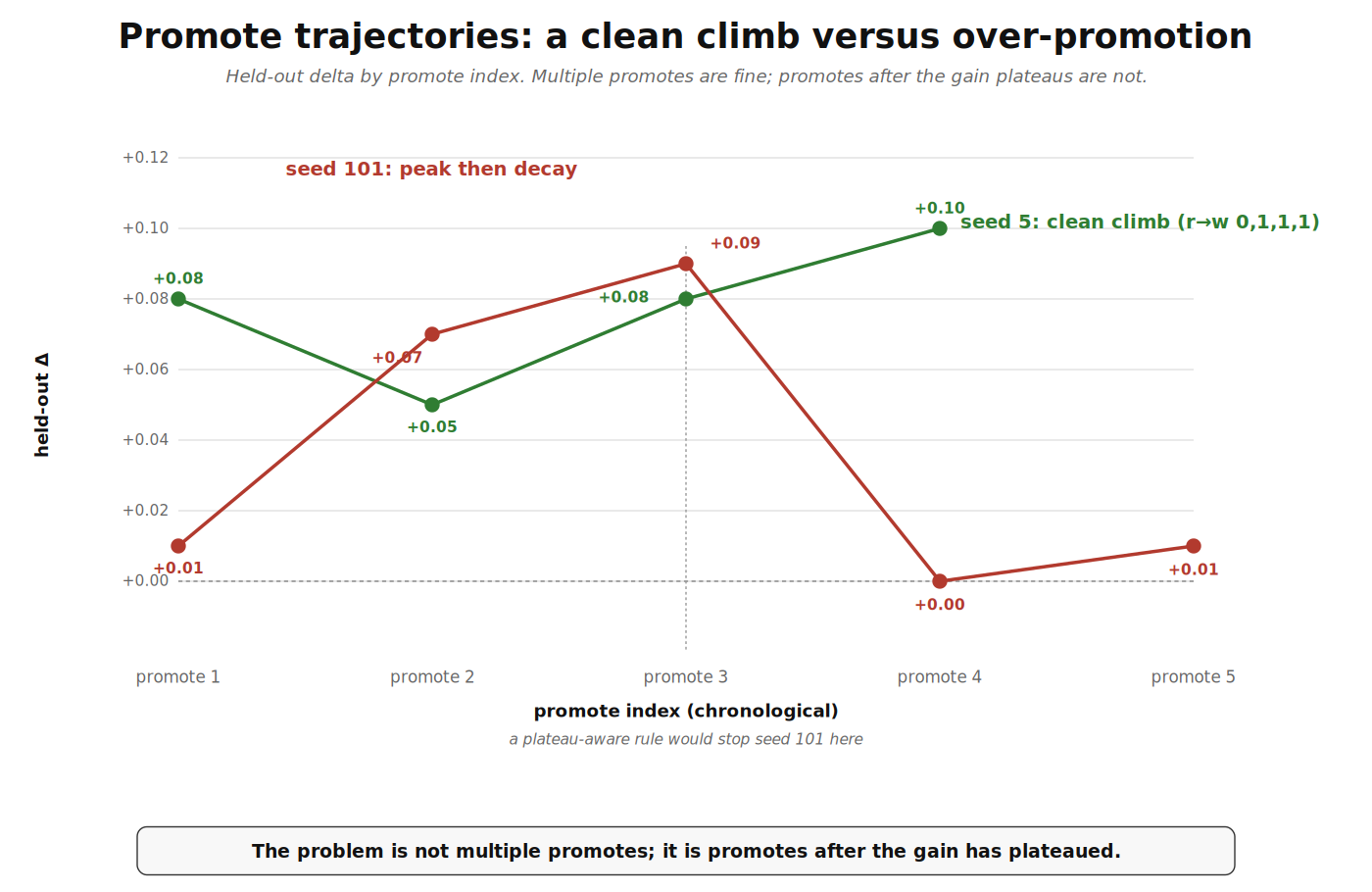}
\caption{Promote trajectories for seed 5 and seed 101, held-out delta by promote index. Seed 5 climbs cleanly (+0.08, +0.05, +0.08, +0.10) with right-to-wrong flips staying at zero or one. Seed 101 peaks at its third promote (+0.09) then decays to +0.01 as later promotes add text without accuracy. The contrast is the basis for the plateau-aware stopping rule proposed in Section 8.2: the problem is not multiple promotes, it is promotes after the gain has plateaued.}
\label{fig:trajectory}
\end{figure}

\subsection{Mechanism: the seam mapping behaves as designed}

Across the five fresh splits, the reader-prompt seam (targeting assemble-internal) helped in every split, producing the positive held-out deltas in Table 1b. The score-transform seam (targeting budget-truncation) tells a sharper story, consistent with the regime-to-seam mapping.

Score-transform accounting across all splits:

\begin{center}\small
\begin{tabular}{ll}
\toprule
score-transform candidates (all splits) & count \\
\midrule
authored & 13 \\
static-rejected & 1 \\
discarded on OPTIMIZE & 9 \\
discarded on CONFIRM & 2 \\
promoted & 1 \\
worst multi-session regression among discarded & minus 0.46 \\
\bottomrule
\end{tabular}
\end{center}

Twelve of thirteen score-transforms were discarded, several regressing multi-session accuracy by up to 0.46. The single promoted score-transform was in seed 101, where budget-truncation tied assemble-internal as the dominant failure regime (three budget-truncation failures of six total, the highest budget-truncation share of any split), and there it produced a real held-out gain: CONFIRM 0.80 to 0.85, +0.05, with 6 wrong-to-right and 1 right-to-wrong. One caveat matters for interpreting this gain: it was an intermediate state, not the final one. Seed 101's subsequent reader-prompt over-promotion (Section 5.7) eroded the cumulative accuracy, and the final deployed seed-101 state was 0.79, below the 0.85 the score-transform reached, so the score-transform's marginal success is real but its benefit did not survive the later drift. In the splits where budget-truncation was sparse relative to assemble-internal, score-transforms targeting it found no generalizable signal and were discarded.

Two things must be kept separate here. Routing is share-based by construction: the loop drafts a transform for the dominant regime, the argmax of the failure histogram, so which seam gets attempted is a relative comparison within a split, not an absolute-count threshold. That is a property of the mechanism and holds across all five splits. Whether the drafted transform then survives the held-out gate is a separate question, and the score-transform seam cleared it exactly once, in seed 101, the only split where budget-truncation reached co-dominant share (tied with assemble-internal at three each). In seed 5, budget-truncation had a higher absolute count (seven failures) but was a clear minority against thirteen assemble-internal failures, so it was never the routed target and its score-transforms were discarded. This single positive case is consistent with the regime-to-seam mapping but does not establish that share, rather than the specific questions or chance, is what let that one transform generalize; with one promotion in five splits we can claim the routing is share-based and that the one generalizing score-transform appeared where its regime was co-dominant, not that share causes seam success. The reader-prompt seam helped broadly because assemble-internal is the dominant live regime in four of five splits, consistent with the paper's larger claim that reconciliation is the dominant wall. budget-truncation is a retrieval-adjacent regime, and one split in five having it reach co-dominant share is consistent with reconciliation being the dominant but not the only wall. The discard pattern is indirect rather than direct evidence for the routing claim: it shows that when a seam is matched to a regime that lacks share the gate correctly promotes nothing, which is the consequence routing predicts, but it does not isolate routing from held-out gating alone. A direct test would ablate the regime labels and let an author work from the failed examples without routing (Threat 11); we have not run that, so the routing attribution rests on this indirect signal.

A second confound on the score-transform evidence deserves to stand on its own, because it cuts at the same claim from a different angle. A score-transform must encode a generalizable continuous reranking function over heterogeneous retrieval lists, while a reader-prompt transform only appends natural language to an instruction. These are different levels of authoring difficulty as intervention classes, independent of regime alignment. So the 12-of-13 score-transform discard rate is consistent with the routing story (the seam was mismatched to regime share) but equally consistent with score-transforms simply being structurally harder to author and generalize. The discard-rate asymmetry cannot distinguish these two explanations without a cross-class authoring-difficulty baseline, which we do not have. Taken together with the missing no-routing ablation above, the honest position is that the routing-as-mechanism claim is supported only indirectly and is the paper's primary open question.

The promoted reader-prompt transforms (full text, Appendix A) were drafted independently with no shared state and converge on the same reconciliation moves: trust retrieved context (anti-hedging), count and synthesize across all sessions, and resolve relative time against session dates. The knowledge-update-heavy seed 7 split additionally surfaced supersession, which is the rule to use the later value when two sessions conflict, discovered precisely when the split contained the failures that need it. That independently drafted transforms rediscover the same principles is evidence the held-out gain is a real mechanism rather than a lucky prompt.

The difference between transforms is equally informative. The lower-baseline runs carried question-type-specific clauses, for example literal "how many weddings" and "sale price versus original price" examples (flagged in Appendix A), and those are the clauses that fail to generalize. The held-out set washes out the overfit and leaves the general core.

\subsection{The regressions as the strongest signal}

Seed 7's two held-out regressions are the most instructive data here. Both were answered correctly at baseline and fail for the same reason, which is a type-conditional prose rule firing on a question where it does not apply.

gpt4\_e414231f, "Which bike did I fix or service the past weekend?", gold answer road bike, typed temporal-reasoning. This is temporal retrieval, finding the event in the right window, not date arithmetic. The transform's temporal rule, which is to subtract a birth year from an event year to find an age when both are present, pushed the reader toward computation the question never required.

618f13b2, "How many times have I worn my new black Converse?", gold answer six, typed knowledge-update. The task is counting, but the type label is knowledge-update, so the transform's supersession rule fired, which says to always use the most recent value for knowledge-update. For a counting question this is wrong, because the answer needs summation and the latest value returns a single number, not six.

The shared root cause is that the rules condition on the coarse question-type label, and the label is a lossy proxy for the reasoning the question actually needs. Prose cannot condition on "only when this applies." It fires whenever the type matches. This is the empirical argument for Section 8: the promoted prompts are probes whose successes name a useful evidence-use behavior and whose regressions name the structure on which that behavior should be conditioned.

\section{What the held-out discipline bought, the classifier, determinism, and a reporting gap}

The strict OPTIMIZE and CONFIRM split with a binding held-out gate is the decision that made this produce trustworthy outputs rather than plausible noise. It worked three ways. It rejected most score-transforms that improved in-sample but regressed on held-out (Section 5.8). It passed the reader-prompt transforms at the deltas reported in Table 1b. And it discounted overfit within a promoted transform, where one run's in-sample +0.18 became held-out +0.04. The one place the gate is too weak is its default threshold of 0.0, which permits over-promotion at high baselines (Section 5.7); the threshold is configurable and the fix is built on that hook (Section 8.2).

Viewed as a scatter of OPTIMIZE delta against CONFIRM delta with a non-regression boundary at zero, the reader-prompt transforms sit in the promotable region, the discarded score-transforms sit below the line, and the single promoted seed-101 budget-truncation score-transform sits above it. Most score-transforms are rejected, and the one that generalized did so in the split where its regime had mass.

Failure-regime detection. The regime classifier is a deterministic oracle over gold evidence locations, not an LLM judge, embedding similarity, or lexical heuristic. For each failed question it reads the benchmark's gold session IDs and, when available, gold evidence-turn IDs, together with the per-turn score dict and the assembly decisions, and assigns one of: scoring-error (no gold turn surfaced in the scores), assemble-internal (gold retrieved and assembled but the answer is still wrong, the reconciliation case), budget-truncation (well-ranked gold evidence dropped at the context budget), or assembly-crowding (gold ranked well but not selected). The exact predicate thresholds, the evidence-turn versus session-level granularity, and a fixed mislabeling bug (commit eac54add, which had let a far-ranked evidence turn masquerade as budget-truncation when it was really a signal gap) are documented in Appendix B.

Two consequences matter for the claims. First, the classifier consumes gold evidence locations, not gold answers; question correctness is scored separately by the unmodified LongMemEval judge (the benchmark's LLM-as-judge comparing the reader's answer to the gold answer; \citet{wu2024longmemeval}), so the diagnosis layer and the correctness signal are independent. Because that judge is an LLM scoring answer content, a confound is worth naming: the anti-hedging clause that opens every promoted transform could convert a hedged "I cannot find" into a confident direct answer, and if the judge rewards directness independently of correctness, some wrong-to-right flips would reflect format compliance rather than reconciliation (Threat 10). The abstention check (no observed abstention cost, Section 5.6) bounds a different failure mode, namely false answers when evidence is genuinely absent; it does not bound this confound, because the wrong-to-right flips that carry the result are on evidence-present questions, where a judge rewarding directness would inflate the count regardless of abstention behavior. Second, because the classifier consults gold evidence, it is a measurement instrument available offline for diagnosis, not a signal the agent could compute on unlabeled deployment traffic. This is deliberate: the diagnostic layer is an oracle kept separate from the agent it diagnoses (Section 9), in the same spirit as a held-out label set. The open item is not the mechanism, which is fully specified, but human validation that these oracle labels match human judgment on a sample (Table 2, Section 8.1).

\textit{Table 2 (regime-classifier human-validation audit, proposed): a small manual audit, columns regime, sampled n, human-agreement, common confusion. The classifier mechanism is fully specified above; this table is the remaining validation step, human-labeling a sample of seed 7 held-out flips and regressions to estimate how well the oracle labels match human judgment, especially assemble-internal versus budget-truncation. A question-type by failure-regime heatmap can present the cross-tabulation once the sample is labeled.}

Determinism claim, narrowed but strengthened by the audit. The regime classifier is deterministic oracle computation over gold evidence locations, so the diagnosis layer introduces no model nondeterminism; the only nondeterministic components are the reader and author model calls. The ActiveGraph substrate guarantees byte-reproducible replay of a run from its event log, including a content-addressed cache that records model and tool responses so replay performs no new model calls \citep{nakajima2026log}. We therefore claim deterministic replay of the non-LLM pipeline (including diagnosis) plus cached model responses, meaning a recorded run replays byte-identically. We do not claim deterministic regeneration of model outputs from fresh API calls. As noted in Section 5.1, even the greedy temperature-0.0 reader can vary slightly across fresh calls because of hosted-model nondeterminism (Threat 2). Our reproducibility rests on cached responses and seeded splits, not on the hosted model returning identical text on a new call.

A scope note on what replay verifies. The byte-identical replay guarantee is substrate-internal: it reproduces a recorded run exactly within an environment that holds the content-addressed response cache. The committed artifacts include the per-question outcome files and analysis scripts but not the response cache itself, so an external researcher can re-derive every reported statistic from the committed per-question outcomes, but cannot independently regenerate those outcomes by replay without access to the cache (which can be released on request) or fresh model calls (which the substrate does not promise will match). We state this rather than imply that committing the scripts alone makes the model outputs externally reproducible.

A reporting gap worth recording. The two earlier fixed-split runs saved only the aggregate held-out delta, not per-question outcomes. The detailed tables and significance could be computed only after the report writer was fixed to persist per-question CONFIRM outcomes plus bidirectional attribution. That fix also added the split-seed flag used for the five fresh splits. Persist per-instance evaluation data, not just aggregates, because a confusion table cannot be reconstructed from a scalar. This gap also qualifies the auditability claim for those two runs specifically: the missing per-question outcomes are not recoverable by re-replaying their event logs, which means the evaluation outputs in those runs were not written to the log as events, so the auditability guarantee did not yet cover the evaluation stage when they ran. The fix persists per-question CONFIRM outcomes as events, closing that coverage gap for every subsequent run; the auditability claim in the abstract and Section 1 should be read as applying to runs on the corrected reporting path, with the two earlier fixed-split runs holding aggregate-only records.

\section{Threats to validity}

The load-bearing threats are surfaced in the main text where they arise (Sections 5.1, 5.5, 5.7, 6); they are consolidated here, followed by an explicit statement of the paper's boundary. The eleven fall into three groups: statistical limitations (1, 2, 3, 7), measurement and mechanism confounds (4, 5, 8, 10, 11), and generalization limits (6, 9). The numbering is retained for cross-references from the main text.

\begin{enumerate}
\item Split-selection sensitivity and shared pool. The five-split replication addresses the single-split concern, but all splits are from one 500-question pool (Section 5.5), so the splits are seeded-independent loop runs, not independent samples from a population. Load-bearing.
\item Reader nondeterminism, roughly plus or minus 0.02 to 0.04, the same order as the fixed-split effects; only the fresh-split effects and the pooled summary clearly exceed it (Section 5.1). Load-bearing.
\item Small held-out n per split, 100, with twelve or fewer discordant pairs per split, so individual splits are underpowered. Small abstention n, 6: no observed abstention cost is reassuring, not conclusive.
\item Regime-classifier labels are oracle-derived and not yet human-validated (Section 6); mislabeling could explain some spillover and the seed-101 score-transform interpretation. Load-bearing, addressed by Table 2.
\item Single reader (claude-sonnet-4-6) and single benchmark, so the improvement is likely reader-dependent and the principles may be LongMemEval-shaped; the paper treats the discovered principles as candidates for reader-independent operators rather than transferable prompts (Section 8.3). The author and reader are the same model, so the author may draft effective reconciliation instructions partly from parametric self-knowledge of the reader's failure modes rather than purely from the regime signal; a structurally different author model would isolate the routing signal's marginal contribution. A preliminary cross-author check addresses this directly: a different author reproduced the reconciliation gain wherever it was strong enough to clear the gate. We re-ran all five splits with claude-haiku-4-5 authoring repairs for the unchanged claude-sonnet-4-6 reader (same splits, same diagnosis, same gate; only the author changed). On the three sonnet-positive splits with power to detect a gain, the haiku author reproduced it within roughly one point and reproduced significance where sonnet had it: seed 5 +0.11 (13 vs 2, p = 0.007; sonnet +0.10), seed 7 +0.09 (11 vs 2, p = 0.023; sonnet +0.08), seed 23 +0.06 (8 vs 2, p = 0.109; sonnet +0.05). On the other two splits the haiku author promoted nothing: on seed 11 (sonnet +0.06) its reader-prompt candidates kept regressing multi-session and failed the gate, a capability limit of the weaker author on an underpowered split rather than evidence against the gain; on seed 101 (six failures, a regime tie, the sonnet over-promotion null) all six candidates failed in-sample. A haiku non-promotion cannot distinguish a real absence of gain from a weaker author missing one, so the positive evidence is the three reproductions, not the two zeros. Because the haiku author does not share the reader's parametric self-knowledge, this is preliminary evidence that the gains are not solely an artifact of shared author and reader identity. It is one alternative author on the same five splits; a structurally different author family and a no-routing ablation (Threat 11) are the controls that would isolate the diagnosis layer's contribution (Section 8.1). Load-bearing.
\item Possible prompt overfitting. Promoted prompts contain some type-specific wording (Appendix A); the held-out gate discounts but may not eliminate benchmark-structure leakage. Earlier fixed-split runs are not auditable per-question.
\item Selection pressure across candidates (Section 5.5). The funnel disclosure bounds this (two CONFIRM-stage discards total), but the McNemar p does not correct for candidates that reached CONFIRM.
\item Over-promotion at high baselines (Section 5.7). The default confirm\_threshold of 0.0 permits near-noise promotions; this is a configuration and stopping-rule gap, not an architectural one, and the threshold is already tunable.
\item Regime-histogram stability at small failure pools. The dominant-regime selection that drives routing is computed from 6 to 21 failures per split, so the ranking carries real sampling variance. In four of five splits assemble-internal dominates clearly (9 of 10, 8 of 13, 11 of 14, 13 of 21); seed 101 is the exception, with budget-truncation and assemble-internal tied at 3 each on the smallest pool (6 failures). The five-split replication gives indirect robustness but does not quantify routing stability, and the seed-101 tie is consistent with that split being the noisiest in baseline, final state, and over-promotion behavior.
\item Correctness judge and format-compliance confound. Correctness is scored by the LongMemEval judge, an LLM-as-judge over the gold answer \citep{wu2024longmemeval}; since the first clause of every promoted transform is anti-hedging, some wrong-to-right flips could reflect the judge rewarding a confident direct answer over a hedge rather than improved reconciliation. The abstention check bounds only the evidence-absent failure mode (no observed abstention cost); it does not bound this confound on the evidence-present flips that carry the headline, which remains unseparated.
\item Routing not isolated by ablation. The claim that regime-directed routing, rather than held-out gating plus failed-example provision alone, drives the reader-prompt gains is not tested against a no-routing control in which an author receives the same failed examples without regime labels or seam constraints. For the dominant assemble-internal pathway these conditions supply similar information, so the attribution of gains specifically to the diagnostic routing is supported only indirectly, by the score-transform discard pattern (Section 5.8) showing that misrouting has visible consequences. A direct reader-prompt ablation is proposed work.
\end{enumerate}

\begin{tcolorbox}[colback=gray!5,colframe=gray!50!black,title=\textbf{What this paper does not show},breakable]
Stated plainly so the boundary is impossible to miss:
\begin{itemize}
\item It does not show that event-sourcing is necessary for an improvement loop. The substrate's demonstrated contributions are auditability (every diagnosis, gate verdict, and promotion is an event in the log) and clean re-homing across targets (the byte-identical extraction, Section 4.1), both shown by construction. We do not compare against a non-event-sourced baseline, so "tractable" and "the lever" are arguments from the affordances the log provides, not a measured advantage over a plain runtime; one could build a target-agnostic loop without event-sourcing and forgo only the auditability and replay guarantees.
\item It does not show cross-task empirical improvement. The second target (text-to-SQL) demonstrates that the loop runs anywhere at the interface and control-flow level; it was not run to a held-out improvement (Section 4.3).
\item It does not show deployment-time regime classification. The regime classifier is an offline oracle over gold evidence locations, usable for benchmark diagnosis, not a signal an agent could compute on live unlabeled traffic (Sections 5.1, 6).
\item It does not show reader-independent transfer. The promoted repairs correct failure modes of one reader (claude-sonnet-4-6) under one prompt and context format (Threat 5).
\item It does not show that guarded operators outperform prose. Section 8.3 proposes operators as designs induced by the failure analysis; none is evaluated here, and distilling a prose principle into a working operator is itself an open problem (Section 8.3).
\item It does not establish the pooled effect as independent-sample evidence. All splits come from the same 500-question pool; the pooled count is descriptive (Section 5.5).
\end{itemize}
\end{tcolorbox}

The defensible result is narrow and stated as such: a directionally consistent, modest held-out improvement on one benchmark under one reader, two of five splits individually significant, with the durable contributions being the gated loop, the auditable substrate, and the operator thesis (proposed, not yet evaluated).

\section{Discussion: from prompt transforms to guarded operators}

\subsection{Strengthening the empirical claim}
The five-split replication is complete and reported here. Further proposed work, in rough priority order: human validation of regime labels on a sample of seed-7 flips and regressions (Table 2), which closes the single weakest link, since the localization and seam-mass claims rest on those labels; two controls that would isolate what the diagnostic architecture contributes, namely a no-routing ablation (an author given the failed examples and the held-out gate but no regime labels or seam constraints, Threat 11) and extending the cross-author check (run on all five splits with a haiku author for the sonnet reader, Threat 5) to a structurally different author family, to test whether convergent transform content survives when the author no longer shares the reader's parametric self-knowledge; a direct noise-band measurement, namely one fresh split with three baseline-and-post repeats, which characterizes the reader noise band with fresh API calls rather than inferring it from cross-run variation (cached replay cannot do this, Section 5.1); a clean significance test that re-evaluates a promoted pipeline on a third split withheld from both OPTIMIZE and CONFIRM, which removes the sequential-selection pressure that the marginal gate reduces but does not eliminate on multi-promote splits (Section 5.5); an abstention stress set, with cases of absent, misleading, stale, and contradictory evidence, which are the failure modes the anti-hedging instruction most endangers; a full-500 evaluation of a promoted transform excluding the OPTIMIZE questions it was tuned on, to estimate benchmark-level effect without optimization leakage; and a dedicated investigation of operator distillation (Section 8.3), turning the proposed operators into working guarded operators with learned or model-backed extraction and evaluating them on held-out splits.

\subsection{A held-out-plateau stopping criterion}
Seed 101 (Section 5.7) shows the default confirm\_threshold of 0.0 over-promotes at high baselines, and the seed-5 contrast shows that multiple promotes are fine as long as they keep producing real gains. The code already exposes confirm\_threshold as a tunable (Section 5.1), so part of the fix is a configuration change: default it above zero, to roughly +0.02, to require clearing the reader noise band. The remaining and more interesting part is a plateau rule layered on top: halt promotion when successive promotes stop producing flip-count gains beyond the noise band, for example requiring the net flips of a new promote to exceed a threshold calibrated to the per-split discordant-pair noise rather than merely a positive accuracy delta. Predicted effect: seed 101 would stop after its third promote (the +0.09 state) instead of drifting to +0.01, while seed 5's clean four-promote climb would be preserved. Both changes are concrete and testable, and the threshold half is already a one-line config.

\subsection{Operator design sketch (the central forward bet)}

The loop reliably rediscovers a small set of reconciliation principles (Section 5.8), and its failures come from those principles, expressed as prose, firing where they do not apply (Section 5.9). The thesis is that these principles should be promoted from prose into guarded deterministic operators that fire on detected structure rather than coarse type labels. The prose transform overfires because prose cannot condition; a guarded operator fires precisely because it conditions on detected structure. The two seed-7 regressions in Section 5.9 are the empirical signal for this thesis. The following operators are design hypotheses induced by the prompt-transform failures; they are proposed here and not evaluated in this paper.

\begin{itemize}
\item count\_entities\_across\_sessions. Fixes accumulation and "how many times" questions; detector fires when the question carries counting intent and the matching evidence is a set of distinct events across sessions, returning their sum.
\item select\_current\_fact\_under\_supersession. Improves changed-fact questions without hurting counting, by firing only when the same attribute is observed with different values across dated sessions, which is a detected change, not whenever a question is labeled knowledge-update. This is the complement of the counting operator: together they would split the single prose supersession clause that caused the Converse regression into a count path and a change path that are mutually exclusive by guard.
\item verify\_evidence\_absence\_before\_abstention. Preserves abstention accuracy while permitting stronger context-trust; the guard that makes anti-hedging safe by checking for genuine evidence absence before suppressing a hedge.
\item resolve\_relative\_time\_against\_session\_date. Helps relative-date questions without inducing unnecessary arithmetic; computes dates from session timestamps but performs age or duration arithmetic only when the question requests a quantity; would not have triggered age-subtraction on the bike-service question gpt4\_e414231f.
\item fallback\_to\_raw\_span. When synthesis confidence is low, return the supporting span rather than an over-synthesized answer.
\end{itemize}

Converting these prompt-discovered principles into guarded operators is itself a nontrivial research step, not a mechanical translation. The prose transform offloads extraction onto the reader's comprehension: when it works, it works because the reader understands which retrieved quantity is the current total and which event is distinct from which restatement. A deterministic operator must replace that comprehension with explicit detection, and a lexical proxy is a poor substitute. We therefore treat operator distillation as its own investigation, likely requiring learned or model-backed extraction inside the guard. Open questions include whether operators are hand-coded, learned, or synthesized by Regimes itself, and how a detector distinguishes a changed fact from an accumulating count.

\subsection{Connection to the substrate}

The prompt transform is the discovery mechanism, not the destination. Regimes reveals which evidence-use behaviors matter by finding them under held-out pressure, and those behaviors should then be promoted into deterministic graph operators that fire on detected structure. This connects to the event-sourced substrate \citep{nakajima2026log}: semantic facts should not replace the event log; they should be typed projections that route back to source turns. Once failures are typed against that architecture, Regimes decides which seam is worth changing, and the reconciliation behaviors it discovers are the operators that belong in the assembly and reader path, conditioned on the graph's own structure rather than on a question's coarse label. A natural synthesis with the concurrent GRASP work (Section 2), left to future work, is a system that gates operator proposals the way GRASP gates skill edits, using regime-to-seam routing to decide which operator class to propose.

\section{Design notes that generalized}

\begin{itemize}
\item Taxonomy is a measurement instrument, and the loop does not rewrite it. New regimes can be registered under validation, but Regimes improves behavior against a fixed diagnostic layer, since letting the optimizer modify its own measurement would let it move its own goalposts.
\item No mechanism yet prevents prompt cruft from accumulating across many promoted transforms. Seed 101 is the empirical face of this, where five promotes accumulated text without accumulating accuracy. Many iterations need either a consolidation step, a global prompt-length budget, or the plateau-stopping rule of Section 8.2. GRASP's bounded library with explicit removal is one point in this design space.
\item An autonomous improvement loop is roughly 80 percent measurement and 20 percent search. A model writing improving code sits on a mountain of detector calibration, eval-harness plumbing, persistence, and control-flow hardening. Section 4.4 is what that 80 percent looks like.
\end{itemize}

\section{Conclusion}

The argument of this paper is that the substrate, not the algorithm, is what makes autonomous improvement tractable. An event-sourced agent runtime turns a controlled improvement loop into a first-class workflow: failures are logged, a run replays deterministically from its log, candidate patches scope to typed seams, gates are auditable, and every promote or discard is itself an event. We built one such loop, Regimes, generalized it into a target-agnostic loop and proved it re-homable on a second target, with interface and control-flow generality evidenced by byte-identical event logs, and the act of proving it surfaced and fixed hidden couplings. A prior result on the same substrate named a wall, which is reconciliation rather than retrieval. Given an action space that reaches the wall, the loop produced a modest held-out improvement that is directionally consistent across five seeded fresh splits, four clearly positive at +0.05 to +0.10 and one near zero from over-promotion, with two splits individually significant and the pooled discordance count significant only under a descriptive same-pool summary, and with no observed abstention cost and no observed per-type regression on the fully audited split. The reader-prompt seam helped in every split, and the one score-transform that generalized did so in the single split where its regime reached co-dominant share, consistent with the intended regime-to-seam mapping though resting on that one positive case. The loop rediscovered the same reconciliation principles across independent runs, and its mistakes named its successor, which is guarded deterministic operators that fire on detected structure rather than coarse labels. The effect is small. The durable contributions are ActiveGraph as an auditable substrate that makes controlled improvement loops tractable, the reproducible held-out-gated loop it supports, the binding held-out gate that separated real gains from overfit, the regime-to-seam taxonomy as an organizing design pattern whose marginal contribution is the primary open question, the multi-seed evidence including the over-promotion finding that localizes a concrete stopping-criterion fix, and the hypothesis that the promoted prompt points past itself to a guarded operator. Concurrent work (GRASP) independently validates the one component our loop shares with it, that the gate separates real improvement from overfit, in a different domain, which is supporting evidence that the shared component generalizes even where our own empirical claim does not yet reach.

\appendix
\section{Promoted transform summaries, with overfit annotations, and one full transform}

All promoted reader-prompt transforms are appended to the reader's instruction fragment. General means principles recurring across runs. Type-specific means clauses naming a question type or concrete example, flagged as overfit risk. The summaries below characterize each promoted transform; the full source of every transform is committed under \path{results/run_seed*/promoted_reader_prompt_transform.py} and \path{results/run_2026-05-31_seed7/}. We reproduce the seed-7 transform in full below, because seed 7 is the audited split and the source of the two analyzed regressions.

Full seed-7 (Run 3) promoted reader-prompt transform (the text appended to the reader instruction; this is the committed transform, with the two overfitting clauses called out inline by the annotations in brackets, which are not part of the transform text):

\begin{verbatim}
CRITICAL RECONCILIATION GUIDANCE:
1. Trust the retrieved context. If the context contains the
   information, answer directly. Do not respond "I cannot find"
   when the evidence is present.
2. The sessions are dated. Anchor every time reference to the
   session date, and resolve relative expressions ("last week",
   "the past weekend") against that date.
3. When the answer depends on multiple sessions, count and
   combine across all of them rather than answering from a
   single session.
4. For knowledge-update questions, ALWAYS use the MOST RECENT
   value when a fact changes across sessions.
   [overfit: supersession rule keyed on the knowledge-update
   TYPE LABEL; misfired on "how many times have I worn..."
   which is labeled knowledge-update but needs a COUNT, not
   the latest value. See Section 5.9.]
5. For age or duration questions, subtract the birth year from
   the event year to find the age at that event.
   [overfit: arithmetic rule that fired on a temporal-RETRIEVAL
   question ("which bike did I service") that needed no
   arithmetic. See Section 5.9.]
6. Synthesize across the context; do not stop at the first
   matching turn.
7. Answer directly and specifically.
\end{verbatim}

The general clauses (1, 2, 3, 6, 7) recur across all promoted transforms. The two bracketed clauses (4, 5) are the type-conditional rules whose misfires produced seed 7's two held-out regressions; they condition on the coarse question-type label rather than detected structure, which is the empirical argument for guarded operators in Section 8.3.

Core principles versus overfit clauses across the promoted transforms (the general core recurs; the overfit clauses are split-specific and are what fail to generalize):

\begin{center}\small
\begin{tabular}{p{0.16\textwidth}p{0.40\textwidth}p{0.36\textwidth}}
\toprule
transform & core principles (general) & overfit clauses (split-specific) \\
\midrule
Run 1 (fixed split) & trust-context, count across sessions, date-anchor, synthesize, direct answers & ``how many weddings'' example; ``sale price vs original price'' example; per-type framing \\
Run 2 (fixed split) & session-date timeline, relative-time resolution, count across sessions, anti-hedging, synthesize & a gift/detail example (leaner than Run 1) \\
seed 7 (Run 3) & trust-context, date-anchor, count and combine, synthesize, direct answers & supersession-by-type-label (clause 4); birth-year age arithmetic (clause 5) \\
seeds 11, 23, 5, 101 & same general reconciliation core & minor per-split phrasing; seed 101 accumulates text without accuracy past its plateau \\
\bottomrule
\end{tabular}
\end{center}

The full source of every promoted transform is committed (Appendix B). The per-promote held-out deltas for the multi-promote splits are in Table 1b and Section 5.7 (seed 5: +0.08, +0.05, +0.08, +0.10, a clean climb; seed 101: a peak at +0.09 then decay, plus the one promoted budget-truncation score-transform at +0.05, Section 5.8). The two earlier fixed-split runs carried the most type-specific wording and the largest in-sample-to-held-out collapse (Run 1: in-sample +0.18, held-out +0.04), which is the overfit-discounting the gate is designed to catch.

\section{Reproducibility}

All code, committed runs, split definitions, promoted transform sources, and analysis scripts referenced below are available at \url{https://github.com/yoheinakajima/regimes}; the file paths cited in this section are relative to that repository.

Reader: claude-sonnet-4-6 at temperature 0.0, max\_tokens 1024, tool-free (\texttt{src/regimes/eval/real.py}). Author: claude-sonnet-4-6 at temperature 0.2, max\_tokens 2048 (\texttt{src/regimes/loop/hypothesize.py}). Embeddings: OpenAI text-embedding-3-small, with a deterministic hash-embedder fallback for offline replay (\texttt{src/regimes/agent/embedders.py}). Pinned dependency versions: anthropic 0.104.1, openai 2.38.0, numpy 2.4.6, tiktoken 0.13.0, Python 3.11; scipy is not a dependency, and the McNemar exact tests are computed directly via the standard library (math.comb) in scripts/significance.py. Committed to the repository: per-question seed-7 outcomes (baseline and transform, with is\_abstention and regime labels); all five fresh-split reports under results/run\_seed\{5,11,23,101\} and results/run\_2026-05-31\_seed7; the promoted transform sources; the split definitions config/split.seed\{5,7,11,23,101\}.json; the analysis scripts scripts/significance.py and scripts/extract\_confirm\_tables.py; and results/MULTISEED\_FINDINGS.md. The regime classifier is in src/regimes/loop/regimes.py and the held-out promotion gate in src/regimes/loop/behaviors.py. Split construction is seeded and reproducible via scripts/build\_split.py --split-seed N, using a non-default seed so the canonical config/split.json is preserved. A run is reproducible via scripts/run\_loop.py --mode real --full --split-seed N --lme-data <path>. LongMemEval-S data is external to the repository. The earlier fixed-split runs' per-question outcomes were not retained, which is why the detailed tables are computed on the fresh splits.

Regime-classifier predicate detail (Section 6). The two thresholds take the values WELL\_RANKED\_K = 20 (a gold turn is "well-ranked" if it appears in the top 20 by signal score) and ASSEMBLE\_COVERAGE\_FLOOR = 0.5 (at least half of well-ranked gold must reach the selected context). assemble-internal requires gold present in the scores, gold turns ranked in the top WELL\_RANKED\_K, the fraction of well-ranked gold reaching the selected context clearing ASSEMBLE\_COVERAGE\_FLOOR, and an incorrect answer. budget-truncation requires a truncated context, well-ranked evidence, and a gold evidence turn marked \texttt{included=False, reason="budget"} in the assembly decisions. The classifier prefers evidence-turn granularity over session-level whenever the Outcome carries per-turn evidence markers. An earlier session-level version (fixed in commit eac54add) mislabeled a far-ranked gold evidence turn (for example rank 126) that was iterated past during assembly and left a budget-drop trail: the old detector promoted it into the optimizable budget-truncation bucket when it was really retrieval-signal-gap. Requiring well-ranked evidence before assigning budget-truncation corrects this, and it is why the localization tables distinguish evidence-turn from session-level coverage. The classifier is sensitive to WELL\_RANKED\_K: changing it by a few positions can move a borderline gold turn across the well-ranked threshold and reassign its regime, so replication should use K = 20 to reproduce the reported histograms.

Correctness judge (Section 6). Question correctness is scored by LongMemEval's pinned upstream judge, gpt-4o-2024-08-06, run unmodified \citep{wu2024longmemeval}. The judge model is pinned to a dated snapshot rather than a floating gpt-4o alias, so it does not drift across the five split evaluations even though they were run at different times; the cross-run label-variance concern that a moving endpoint would raise does not apply here. Judge calls score new reader outputs and are an evaluation-stage step; the determinism and cache guarantees in Section 6 cover the reader and author calls, and the correctness labels rest on the pinned judge snapshot being stable for a given input.

\bibliographystyle{plainnat}
\bibliography{refs}

\end{document}